\documentclass[Afour,sageh,times]{sagej}
\usepackage{moreverb,url}
\usepackage[colorlinks,bookmarksopen,bookmarksnumbered,citecolor=red,urlcolor=red]{hyperref}
\usepackage{float}
\usepackage[caption = false]{subfig}
\usepackage{xcolor}
\usepackage{listings}
\usepackage[T1]{fontenc}

\begin{document}

\runninghead{De Roovere, Moonen, Michiels and wyffels}
\title{Dataset of Industrial Metal Objects}
\author{Peter De Roovere\affilnum{1}, Steven Moonen\affilnum{2}, Nick Michiels\affilnum{2}, Francis wyffels\affilnum{1}}
\affiliation{\affilnum{1}IDLab-AIRO -- Ghent University -- imec\\\affilnum{2}Hasselt University -- tUL -- Flanders Make, Expertise Centre for Digital Media}
\corrauth{Peter De Roovere, IDLab-AIRO -- Ghent University -- imec, iGent Tower - Technologiepark-Zwijnaarde 126, B-9052 Ghent, Belgium}
\email{peter.deroovere@ugent.be}

\begin{abstract}
    We present a diverse dataset of industrial metal objects. These objects are symmetric, textureless and highly reflective, leading to challenging conditions not captured in existing datasets. Our dataset contains both real-world and synthetic multi-view RGB images with 6D object pose labels. Real-world data is obtained by recording multi-view images of scenes with varying object shapes, materials, carriers, compositions and lighting conditions. This results in over 30,000 images, accurately labelled using a new public tool. Synthetic data is obtained by carefully simulating real-world conditions and varying them in a controlled and realistic way. This leads to over 500,000 synthetic images. The close correspondence between synthetic and real-world data, and controlled variations, will facilitate sim-to-real research. Our dataset's size and challenging nature will facilitate research on various computer vision tasks involving reflective materials. The dataset and accompanying resources are made available on the project website \url{https://pderoovere.github.io/dimo}.
\end{abstract}


\maketitle

\section{Introduction}

Identifying objects and estimating their 6D poses are crucial aspects of many robotics applications \citep{du2021vision}. Manipulating physical objects requires knowledge about their position and orientation. Estimating these poses from image data is a longstanding problem, which remains an active area of research \citep{rad2017bb8, xiang2017posecnn, tekin2018real, li2018deepim, zhou2019objects}.

Existing object pose estimation datasets fail to capture important aspects of real-world industrial use-cases. The Benchmark for 6D Object Pose Estimation (BOP) \citep{hodan2018bop} unifies 12 datasets, of which the majority focus on household objects. These objects are non-symmetric, with rich textures that lead to distinct visual features. In contrast, most industrial objects are symmetric and textureless. Material properties such as colour and reflectance dominate their visual appearance. Other datasets do incorporate textureless industrial objects \citep{hodan2017t, drost2017introducing} but focus on objects with limited reflectivity. However, reflective materials are widespread. Most high precision machine components, for example, are made of shiny metals. Existing depth sensors are unable to capture these parts correctly. 6D pose estimation from RGB images is thus a fundamental problem for reflective objects. This task is challenging, as these objects literally reflect their surroundings, ambiguating their appearance. This leads to widely varying visual appearances. A diverse dataset is needed to develop and evaluate new solutions.

\begin{figure}[ht]
    \caption{Side-by-side comparison of a real-world image (left) and its synthetic counterpart (right).\label{F1}}
    \centering
    \includegraphics[width=\linewidth]{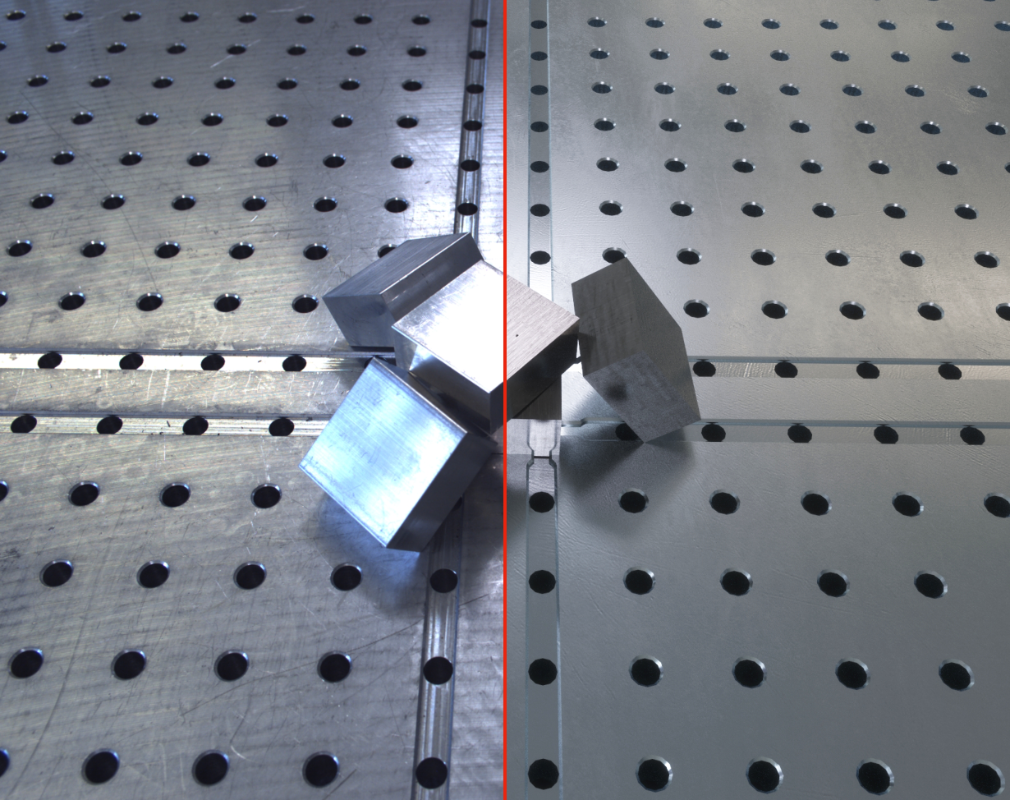}
\end{figure}

As object pose labels are costly to obtain, synthetic data generation is a promising approach to creating diverse pose estimation datasets. Recent advances in computer graphics make it feasible to generate large amounts of labelled data. This has been applied to pose estimation datasets \citep{denninger2019blenderproc, kleeberger2019large}. However, the resulting images fail to capture essential aspects of their real-world counterparts, like object materials, scene compositions, lighting and camera viewpoints. Variations between synthetic and real-world data can significantly hamper model performance -- the so-called \emph{reality gap} \citep{sunderhauf2018limits}. A proposed solution to the reality gap is domain randomization \citep{tobin2017domain}, in which parts of the data generation process are randomized. With enough variability, real-world data may then appear to be just another variation. However, generating \emph{enough} variability to envelope real-world data points is non-trivial. Many variations are wasteful and can unnecessarily increase the task difficulty \citep{hodavn2019photorealistic}.
This problem is amplified for reflective objects, as their visual appearance is highly dependent on their surroundings.

We present a diverse dataset of industrial, reflective objects with real-world data and corresponding synthetic data with controlled variations. Scenes with varying object shapes, materials, carriers, compositions, and lighting are recorded with different cameras and from multiple viewpoints. Our dataset consists of more than 30,000 real-world images labelled using a new open source labeling tool. In addition, we generate over 500,000 synthetic images by carefully mimicking real-world conditions and adding controlled variations in object materials, compositions and lighting. We use modern physically-based rendering techniques to simulate realistic reflections and procedurally generate real-world artefacts like scratches.

\begin{table*}[t]
    \small\centering
    \caption{The different cameras used for capturing real-world image data.\label{T1}}
    \begin{tabular}{llll}
        \toprule
        Camera                 & Lens            & Type                  & Resolution       \\
        \midrule
        JAI GO-5000-PGE        & KOWA LM12HC-V   & RGB                   & $2560\times2048$ \\
        mvBlueFOX3-2124rG-1112 & MV-O1218-10M-KO & Grayscale             & $4064\times3044$ \\
        RealSense L515         &                 & RGB, LiDAR            & $1920\times1080$ \\
        RealSense D415         &                 & RGB, Active IR Stereo & $1920\times1080$ \\
        \bottomrule
    \end{tabular}
\end{table*}

Our dataset will be of value to many different research problems. The close correspondence between synthetic and real-world data, along with controlled variations, will facilitate sim-to-real research. While many existing datasets are insufficiently large \citep{du2021vision}, the large number of labelled images presented in this work can be used for training robust models. Our dataset enables research on many important computer vision tasks involving reflective objects. We focus our discussion on 6D object pose estimation, which is at the root of many robotics use cases. However, our dataset also applies to object detection, instance segmentation, novel view synthesis, 3D reconstruction and active perception.

\section{Methodology}

We built a setup for easily collecting and labelling real-world data and created a system for generating synthetic data by simulating real-world conditions.

\subsection{Real-world data}

\begin{figure}[ht]
    \caption{Different camera sensors are attached to the end-effector of an industrial robot.\label{F3}}
    \centering
    \includegraphics[width=\linewidth]{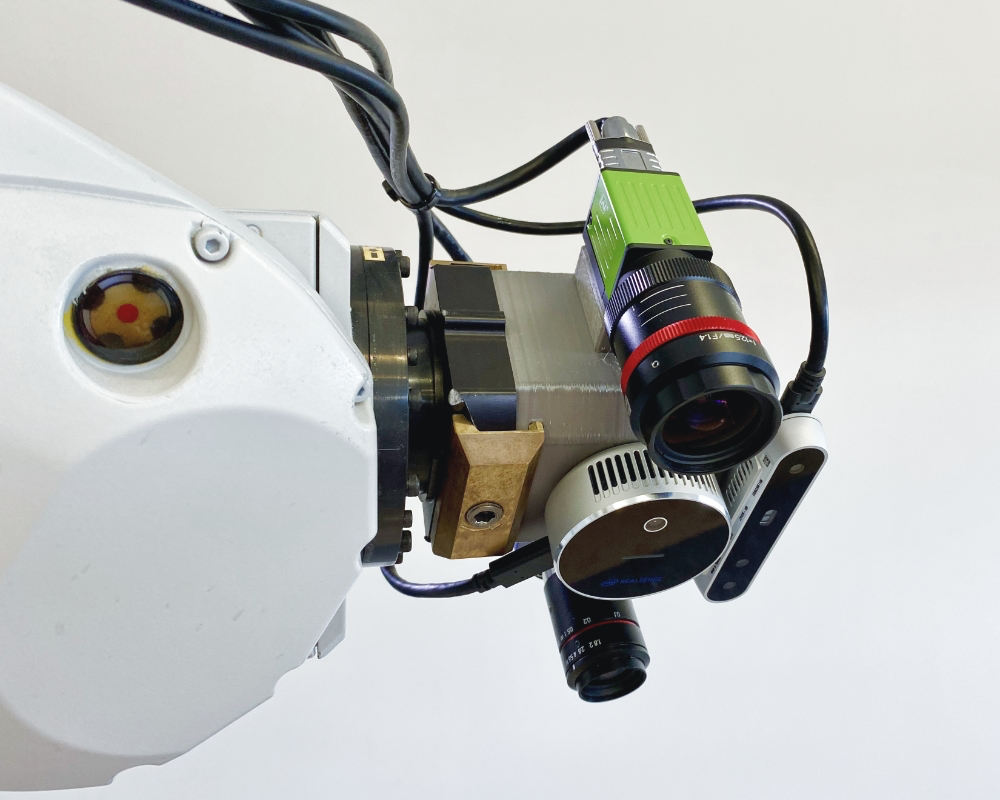}
\end{figure}

\begin{figure}[ht]
    \caption{Top-view visualisation of all 13 camera viewpoints. These viewpoints were spread out on a hemisphere, centred around and oriented towards where objects are placed.\label{F4}}
    \centering
    \includegraphics[width=\linewidth/4*3]{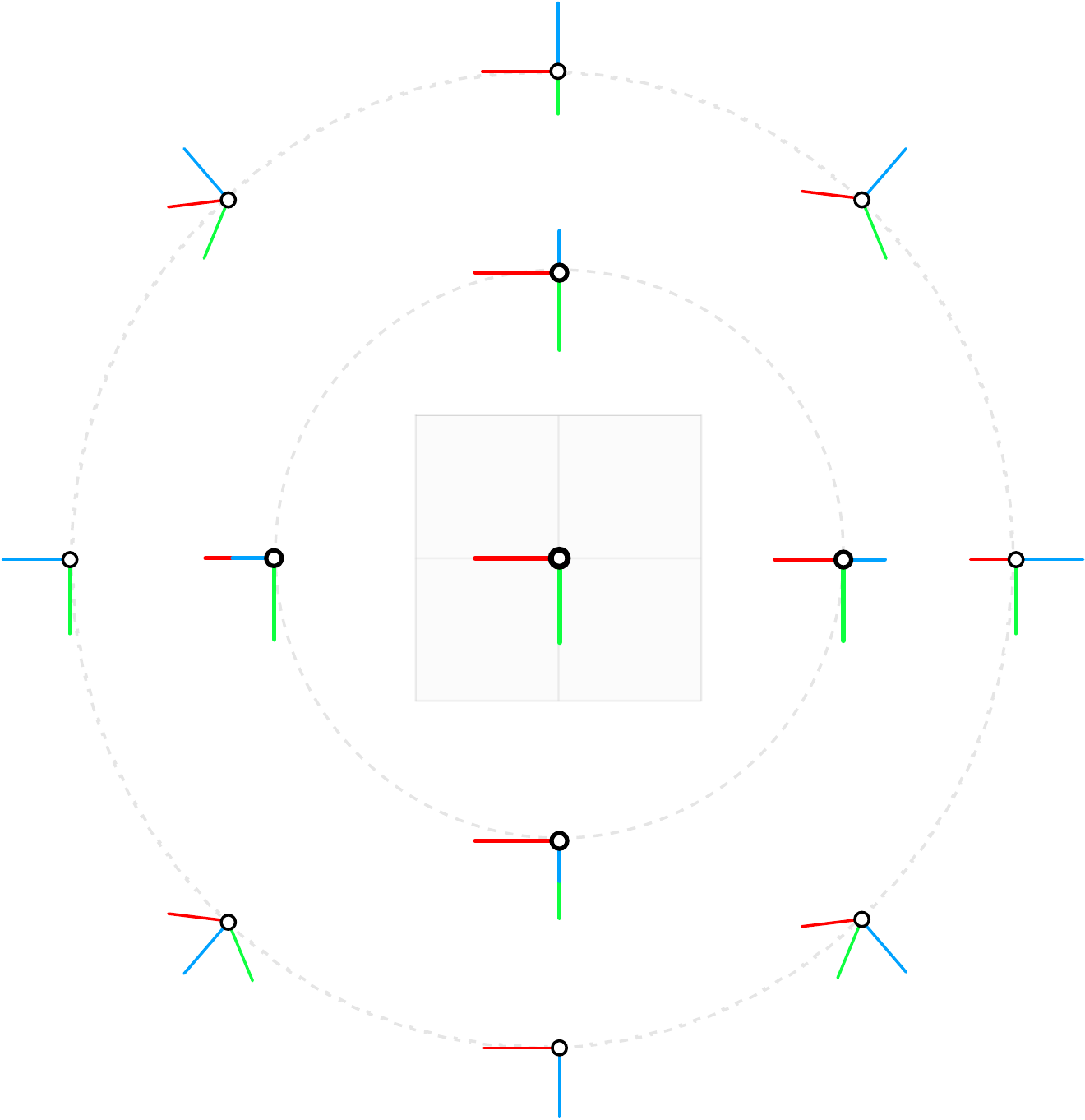}
\end{figure}

We use different camera sensors (Table~\ref{T1}) mounted on an industrial robot, to capture multi-view images of diverse scenes. A custom-designed end-effector is used to attach all cameras to a Fanuc M20ia robot, as shown in Figure~\ref{F3}. The robot, with repeatability of 0.1mm, is used to accurately change camera viewpoints. We further stabilize camera poses by cancelling backlash using the Fanuc \texttt{IRVBKLSH} instruction. We calibrate all cameras using ChArUco targets, undistort all images and perform hand-eye calibration \citep{tsai1989new}. We use high-precision aluminium-LDPE calibration targets with \texttt{DICT\_5x5} dictionary codes and 15mm wide checkers. ChArUco detection and calibration are implemented using OpenCV \citep{opencv_library}. Each scene is captured with every camera from 13 different viewpoints. These viewpoints are spread out on a hemisphere and oriented towards the center of the object carrier, as shown in Figure~\ref{F4}.

\begin{figure}[ht]
    \caption{Our set-up for collecting real-world images. We capture multi-view images of diverse scenes using different camera sensors mounted on an industrial robot.\label{F2}}
    \centering
    \includegraphics[width=\linewidth]{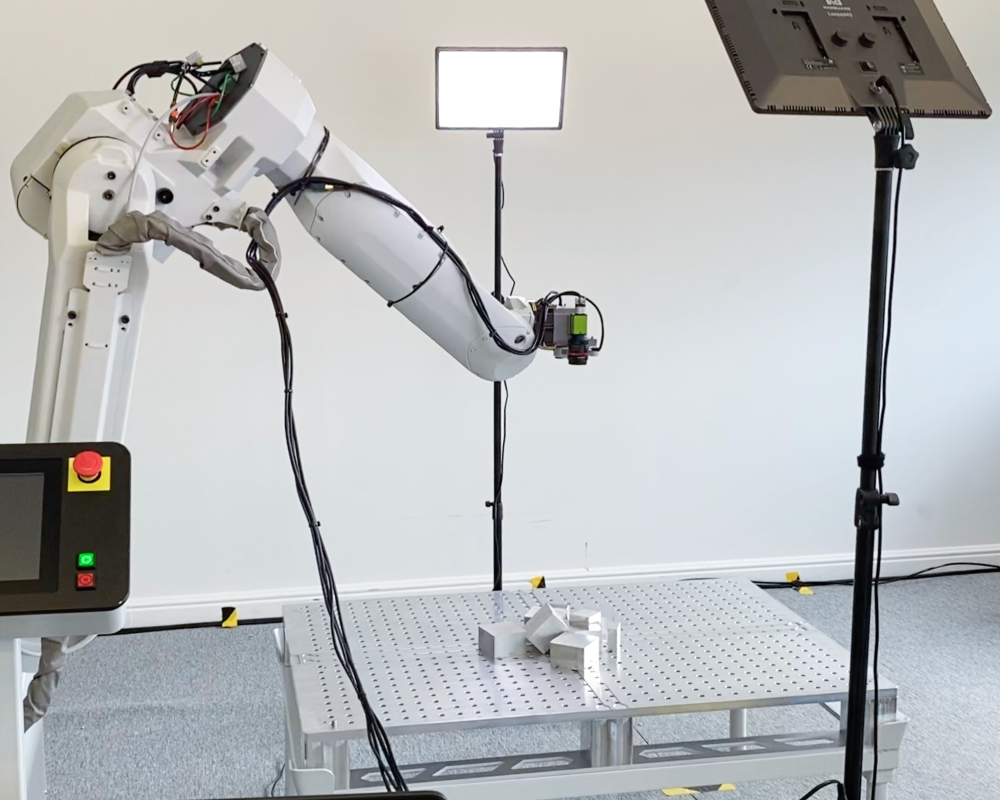}
\end{figure}

\begin{figure*}[t]
    \caption{\textbf{Object categories.} We use objects of 6 categories, with different shapes and material properties. Most objects (\ref{SF1}, \ref{SF2}, \ref{SF3}, \ref{SF4}) are highly reflective and exhibit real-world artefacts like scratches. For comparison, we also include less reflective and textured objects (\ref{SF5}, \ref{SF6}).\label{F5}}
    \subfloat[Cylinder\label{SF1}]{\includegraphics[width=.3\linewidth]{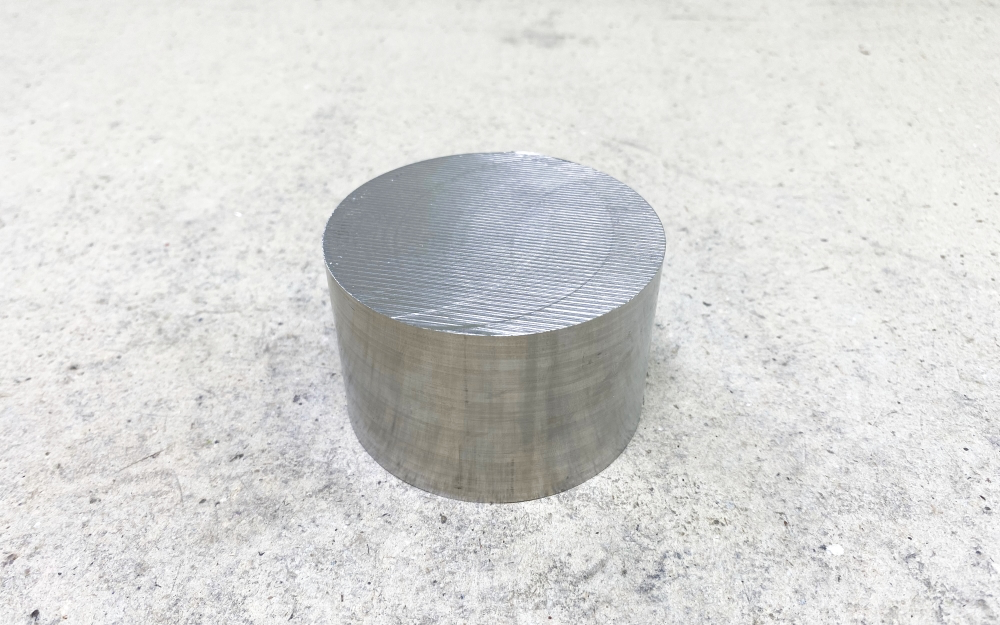}}\hfill
    \subfloat[Small block\label{SF2}]{\includegraphics[width=.3\linewidth]{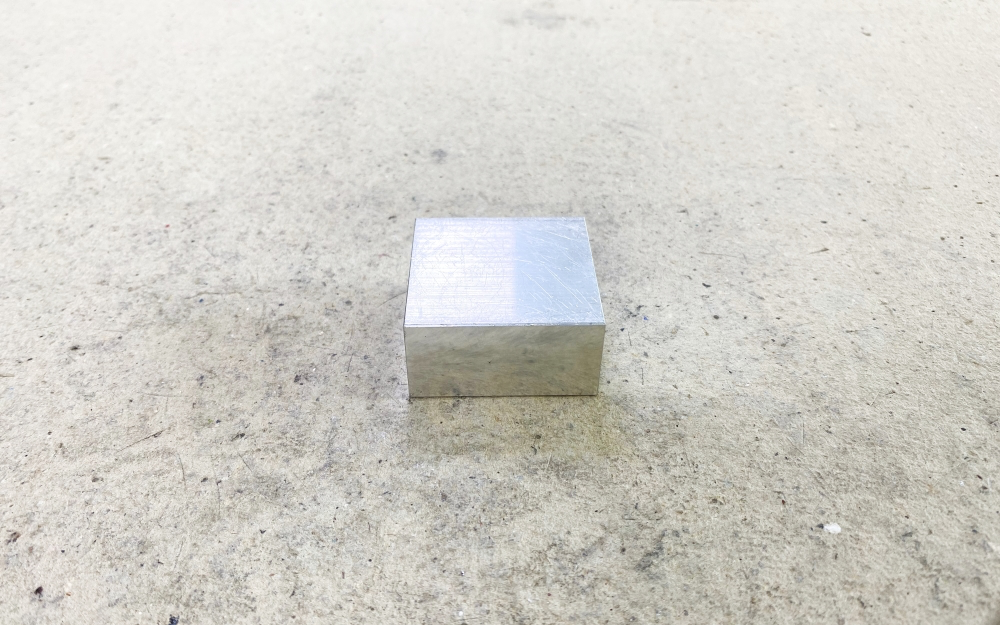}}\hfill
    \subfloat[Large block\label{SF3}]{\includegraphics[width=.3\linewidth]{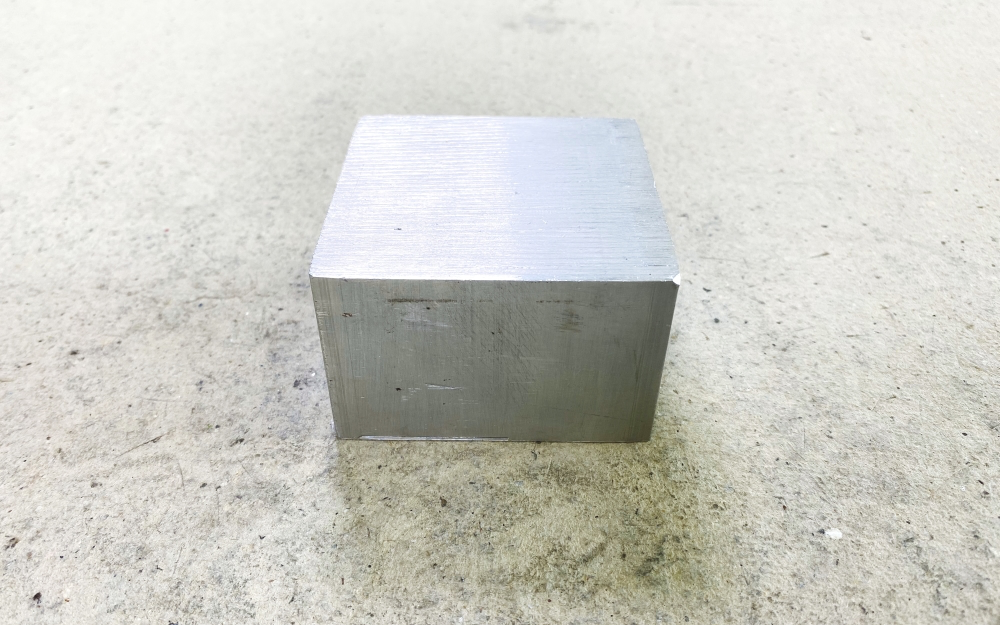}}\\
    \subfloat[Shaft\label{SF4}]{\includegraphics[width=.3\linewidth]{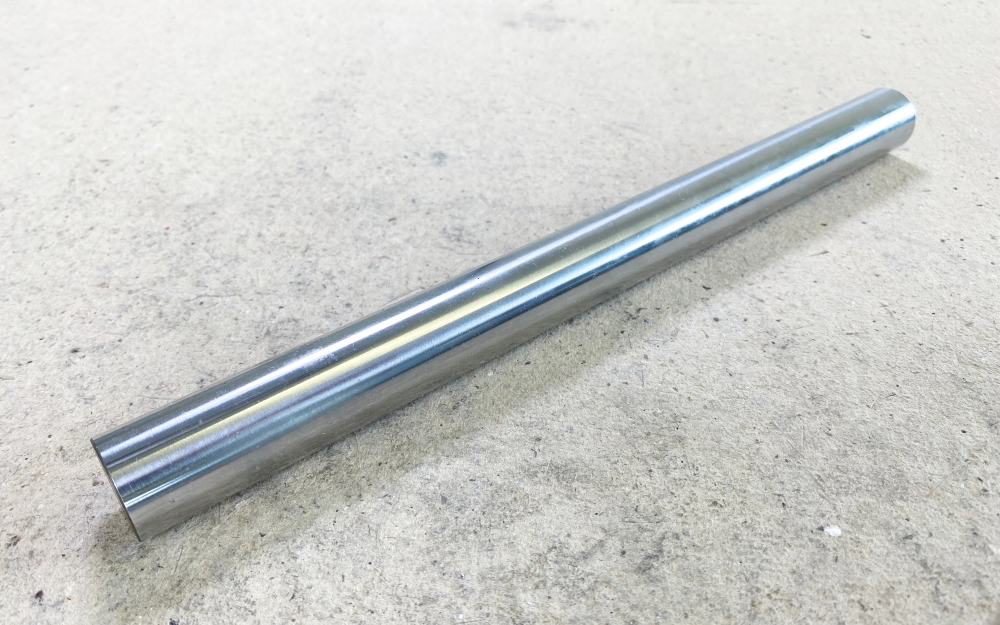}}\hfill
    \subfloat[Red block\label{SF5}]{\includegraphics[width=.3\linewidth]{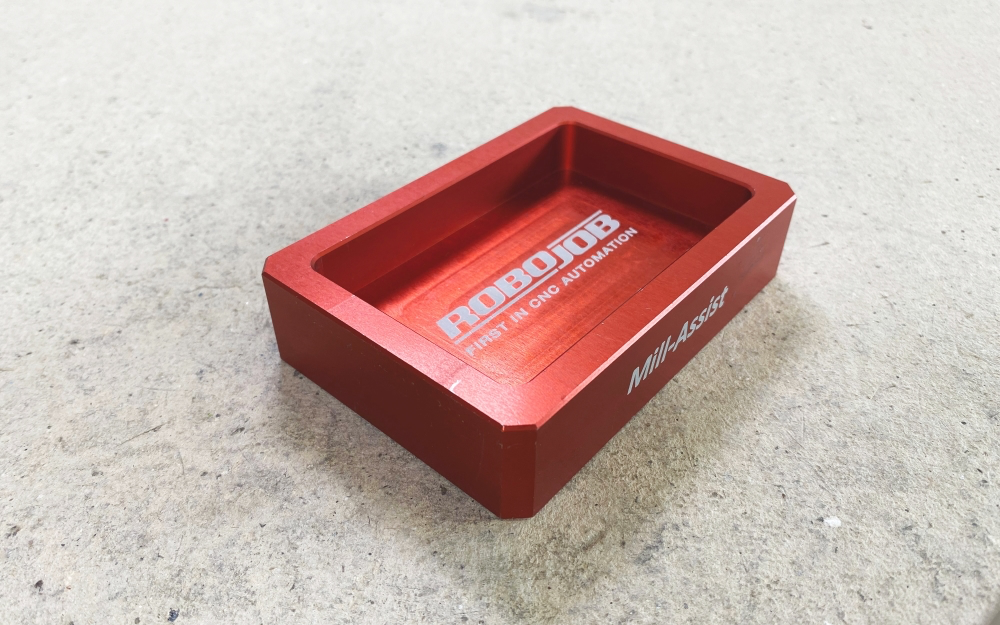}}\hfill
    \subfloat[Red cylinder\label{SF6}]{\includegraphics[width=.3\linewidth]{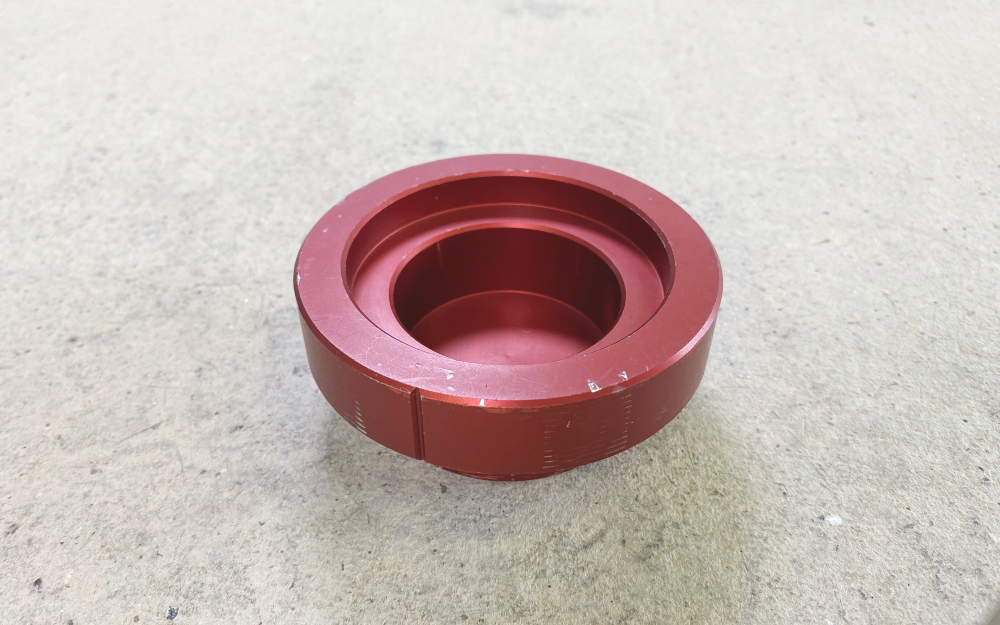}}
\end{figure*}

\begin{figure*}[t]
    \caption{Objects are placed on 3 different types of carriers.\label{F6}}
    \subfloat[Pallet]{\includegraphics[width=.3\linewidth]{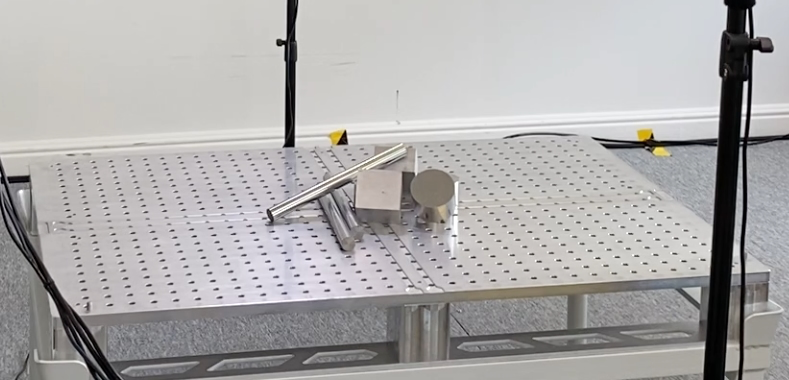}}\hfill
    \subfloat[Bin]{\includegraphics[width=.3\linewidth]{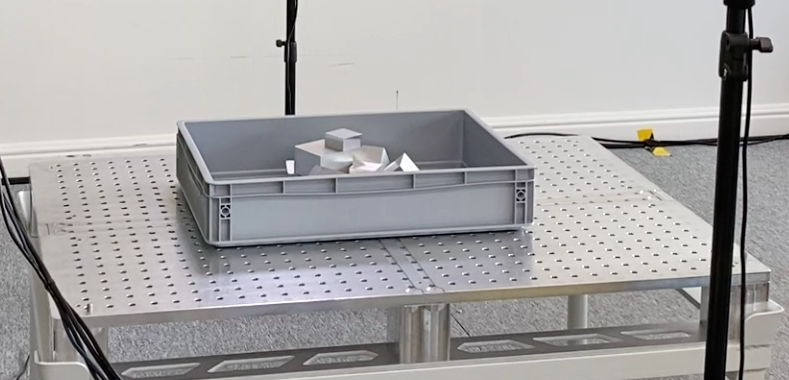}}\hfill
    \subfloat[Cardboard]{\includegraphics[width=.3\linewidth]{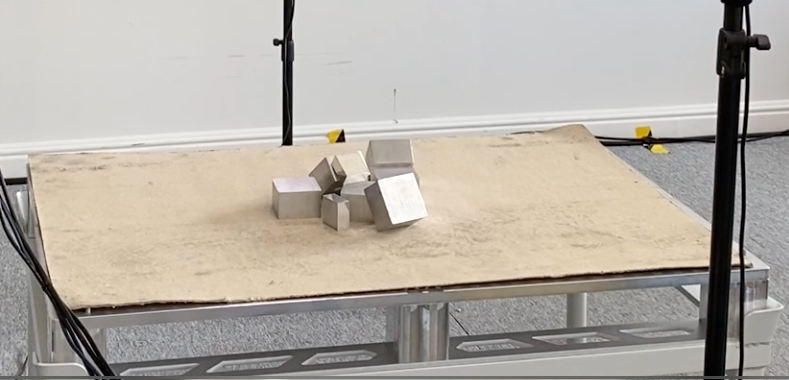}}
\end{figure*}

Diverse scenes are created by varying object shapes, materials, carriers, compositions, and lighting. We use 6 object categories, shown in Figure~\ref{F5}, with different shapes and material properties. Most objects (\ref{SF1}, \ref{SF2}, \ref{SF3}, \ref{SF4}) are highly reflective and exhibit real-world artefacts like scratches and saw patterns. We also include less reflective (\ref{SF5}, \ref{SF6}) and textured objects (\ref{SF5}) for comparison. We use 3 object carriers, shown in Figure~\ref{F6}. Parts are stacked in various compositions with different levels of occlusion. Lighting conditions are varied by controlling Nanguan Luxpad 43 lights. These variations lead to 600 scenes (Table~\ref{T2}) and 31,200 images (Table~\ref{T3}).

\begin{table}[t]
    \small\centering
    \caption{\textbf{Composition variations in real-world data.} We use 6 object types. For each object type, we record 2 compositions with a single instance and 8 compositions with multiple instances (homogeneous). In addition, we create 40 compositions with objects of different types (heterogeneous). This leads to 100 different compositions. \label{T2}}
    \begin{tabular}{ll}
        \toprule
        Category                     & \# Variations         \\
        \midrule
        Object types                 & $6$                   \\
        Single object                & $2$                   \\
        Multiple objects (same type) & $8$                   \\
        \midrule
        Homogeneous                  & $6 \times (2+8) = 60$ \\
        Heterogeneous                & $40$                  \\
        \midrule
        Compositions                 & $60 + 40 = 100$       \\
        \bottomrule
    \end{tabular}
\end{table}

\begin{table}[t]
    \small\centering
    \caption{\textbf{Scene variations in real-world data.} All composition variants are recorded for each carrier (3) and lighting condition (2), resulting in 600 scene variations. Using 4 different cameras, each recording from 13 different viewpoints, this leads to 31,200 images.\label{T3}}
    \begin{tabular}{ll}
        \toprule
        Category     & \# Variations                     \\
        \midrule
        Compositions & $100$                             \\
        Carriers     & $3$                               \\
        Lighting     & $2$ (light or dark)               \\
        \midrule
        Scenes       & $100  \times 3 \times 2 = 600$    \\
        Camera's     & $4$                               \\
        Viewpoints   & $13$                              \\
        \midrule
        Images       & $600 \times 4 \times 13 = 31,200$ \\
        \bottomrule
    \end{tabular}
\end{table}

We carefully label all real-world data using a new open source tool. Precise calibration and consistent camera viewpoints allow joint labelling of all images of one scene. We developed a new open source tool to facilitate this task, available at \url{https://github.com/pderoovere/dimo-labeling}. First, object poses are estimated based on user-annotated correspondences between image pixels and object model points. Next, object models are projected on each image, allowing precise fine-tuning. All poses are defined relative to a global coordinate system, ensuring multi-view consistency. To assess the accuracy of our labelling pipeline, we manually label 10 synthetic scenes, where exact ground-truth poses are available. We measure pose errors using Maximum Symmetry-Aware Surface Distance (MSSD):

\begin{math}
    e_{\text{MSSD}} = \text{min}_{\textbf{S} \in S_O} \text{max}_{\textbf{x}
    \in V_O}
    \Vert \hat{\textbf{P}}\textbf{x} - \bar{\textbf{P}}\textbf{S}\textbf{x}
    \Vert_2,
\end{math}

with $\hat{\textbf{P}}$ the estimated pose, $\bar{\textbf{P}}$ the ground truth pose, $S_O$ a set of symmetry transformations, and $V_O$ a set of vertices of the object CAD model. We obtain an $e_{\text{MSSD}}$ of $0.267\text{mm}$, demonstrating highly accurate manual annotations.

\subsection{Synthetic data}

\begin{figure}[ht]
    \caption{Close up of a synthetic image. Our texture synthesis algorithm leads to realistic object textures, exhibiting real-world artefacts like scratches and saw patterns. The use of path tracing leads to realistic reflections.\label{F7}}
    \centering
    \includegraphics[width=\linewidth]{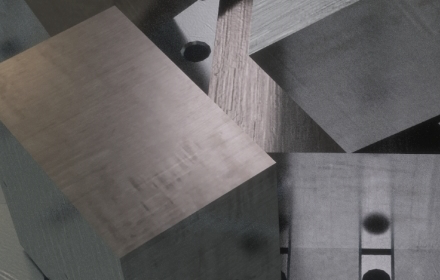}
\end{figure}

Using a Unity project, we generate high-fidelity synthetic data. Real-world conditions are carefully simulated. We adopt camera intrinsics and extrinsics, and model objects and carriers. Environment maps of the different real-world lighting conditions are created by combining bracketed images. We use a Rico~Theta~S~360° camera to capture the environment with different exposures. The resulting images are combined into a single HDRI environment map used to light the virtual scene. We mimic real-world artefacts like scratches and saw patterns by synthesizing textures from real-world images of example objects. These images are relighted and cropped so they can be used as albedo textures. A normal map is created by estimating normal vectors from these textures. A texture synthesis algorithm is then used to create new variations of the captured textures. This texture synthesis algorithm works in two steps. First, random patches are taken from an example image. Next, to reduce seams between copied patches, pixels search their neighbourhood for similar pixels and update their value accordingly. This step is repeated multiple times, reducing the radius of the neighbourhood with each turn, until the neighbourhood consists of a single pixel. This procedure generates unique and realistic appearances for every virtual object. Figure~\ref{F7} shows a close-up of multiple generated object textures. Our synthetic data generation setup leads to synthetic images that strongly resemble their real-world counterparts.

A large amount of additional data is generated by varying object materials, compositions and lighting. Object materials are generated using the texture synthesis algorithm described before, using additional example images. We use Unity’s physics engine to generate physically plausible object poses, starting from random initialisations. Lighting is varied by randomly selecting an environment map from a predefined set. Table~\ref{T4} gives an overview of all variations. Combining these variations leads to 42,600 scenes. To focus our resources on generating environmental variations, we generate synthetic data for 1 camera type (JAI GO-5000-PGE). In total, 553,800 images are generated.

\begin{table}[t]
    \small\centering
    \caption{\textbf{Synthetic variations of a real-world scene.} First, each real-world scene is replicated by mimicking object poses and lighting conditions. Next, lighting conditions are varied while retaining object poses. Next, object poses are varied, while using the original lighting conditions. Finally, both lighting conditions and object poses are jointly varied. Object textures are regenerated for each recording. This leads to 71 synthetic variations for each real-world scene. \label{T4}}
    \begin{tabular}{lll}
        \toprule
        Pose       & Lighting   & \# Variations \\
        \midrule
        Original   & Original   & $1$           \\
        Original   & Randomized & $15$          \\
        Randomized & Original   & $15$          \\
        Randomized & Randomized & $40$          \\
        \midrule
                   &            & 71            \\
        \bottomrule
    \end{tabular}
\end{table}

As we are in total control of the data generation process, labels are free. Generating 1 scene takes 2 minutes using a NVIDIA GeForce 2080 Ti.

\section{Dataset}
Our dataset contains 31,200 images of 600 real-world scenes and 553,800 images of 42,600 synthetic scenes, stored in a unified format. In addition to camera intrinsics and extrinsics, object models and object pose labels, we store information about the data generation process.

\subsection{Format}

\begin{figure}[ht]
    \caption{\textbf{Folder structure of our dataset.} Our dataset contains a subfolder with object CAD models and subfolders for each camera (real and virtual). These camera folders contain subfolders for each scene. Each scene folder contains a subfolder with multi-view images and json files with accompanying information. We adopt the BOP format and extend it (marked in red).\label{F8}}
    \centering
    \includegraphics[width=\linewidth/3*2]{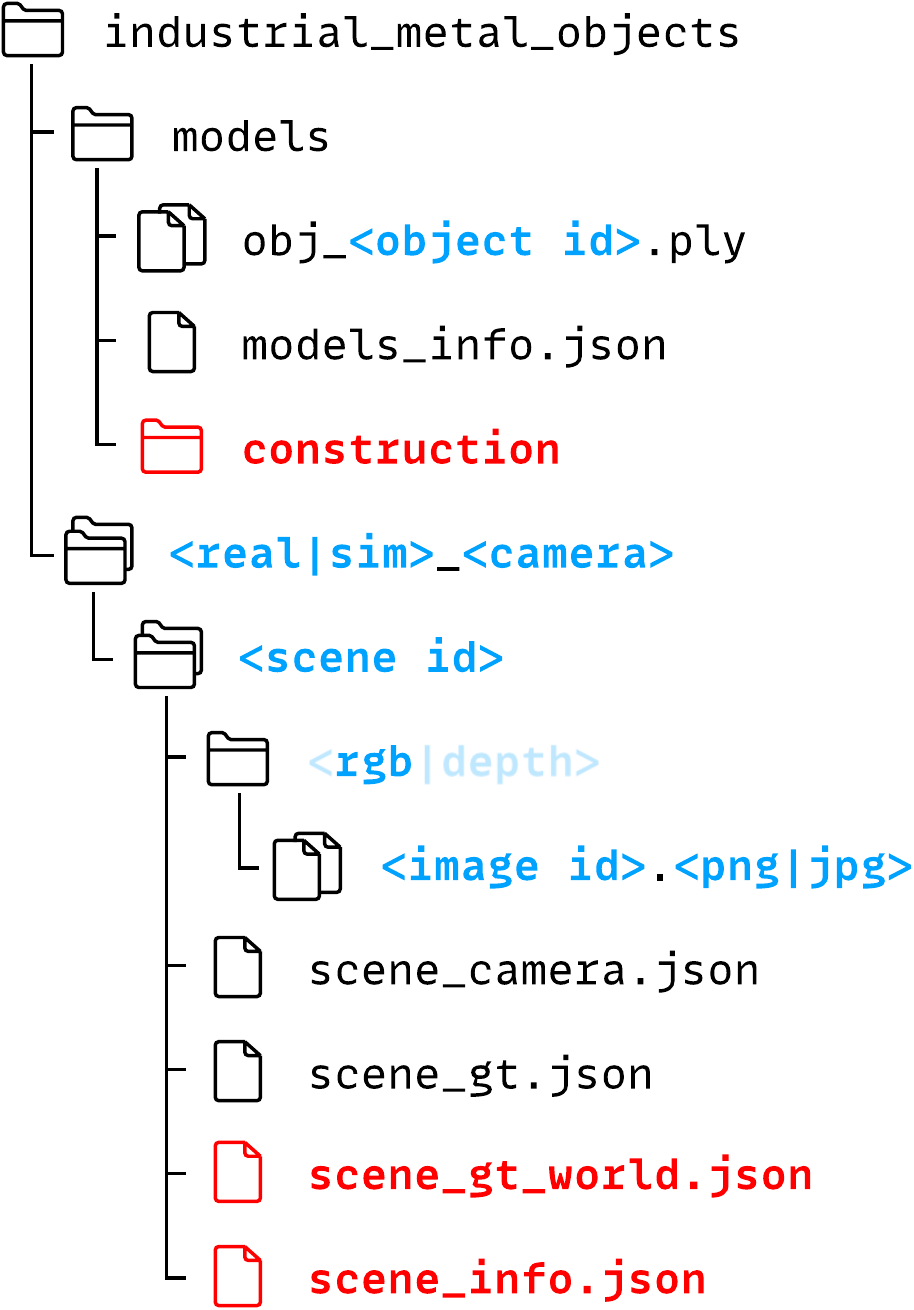}
\end{figure}

We adopt and extend the BOP format \citep{hodan2018bop}. The folder structure is shown in Figure~\ref{F8}. All distances are recorded in mm. Matrices and vectors are stored as flattened lists.

\texttt{models} contains object CAD models, stored in the PLY format. \texttt{models\_info.json} contains additional information about the size and symmetries of each object, as shown in Listing~\ref{L1}. Each continuous symmetry is described by an axis and offset value, each discrete symmetry by a transformation matrix. \texttt{construction} contains technical drawings that can be used for manufacturing the parts.

\lstset{
    string=[s]{<}{>},
    showstringspaces=false,
    stringstyle=\color{cyan}
}

\begin{lstlisting}[caption=\texttt{models\_info.json} contains info about the CAD models.\label{L1}, float]
{
    "<object id>": {
        "diameter": 300.0,
        "min_x": -12.5,
        "min_y": -12.5,
        "min_z": -150.0,
        "size_x": 25.0,
        "size_y": 25.0
        "size_z": 300.0,
        "symmetries_continuous": [
            {
                "axis": <(3,)>, 
                "offset": <(3,)>
            }, ...
        ],
        "symmetries_discrete": [
            <(4,4)>, ...
        ]
    }, ...
}
\end{lstlisting}

Scenes are organised by camera type (both real and virtual) and identified by a unique id. Each scene is available for each camera. For real cameras, 13 images are available for each scene -- one for each viewpoint. For virtual cameras, 71 scenes are available for each real-world scene -- 1 mimicking the corresponding real conditions and 70 generated with varying conditions.

\texttt{scene\_camera.json} contains camera intrinsics and extrinsics for each image, as shown in Listing~\ref{L2}.

\begin{lstlisting}[caption=\texttt{scene\_camera.json} contains camera intrinsics and extrinsics for each image.\label{L2}, float]
{
    "<image id>": {
        "cam_K": <(3,3)>,
        "cam_R_c2w": <(4,4)>,
        "cam_t_c2w": <(3,)>
    }, ...
}
\end{lstlisting}

Object poses are recorded in \texttt{scene\_gt.json} and \texttt{scene\_gt\_world.json}, as shown in Listing~\ref{L3} and Listing~\ref{L4}. For each image, \texttt{scene\_gt.json} contains objects poses relative to the camera reference frame. As poses are consistent between different viewpoints, we also record object poses in a shared global reference frame (\texttt{scene\_gt\_world.json}).

\texttt{scene\_info.json} contains information about the data generation process, as presented in Listing~\ref{L5}. For each image, we record which lighting conditions apply, which carrier is used, the type of composition (heterogeneous or homogeneous) and camera viewpoint. This information allows selecting subsets on the data based on specific variations.

\begin{lstlisting}[caption=\texttt{scene\_gt.json} contains object poses (relative to the camera) for each image.\label{L3}, float]
{
    "<image id>": [{
            "cam_K": <(3,3)>,
            "cam_R_m2c": <(4,4)>,
            "cam_t_m2c": <(3,)>,
            "obj_id": 2,
        }, ...]
}
\end{lstlisting}

\begin{lstlisting}[caption=\texttt{scene\_gt\_world.json} contains object poses (relative to a global world reference). These poses are valid for all images.\label{L4}, float]
{[
    {
        "cam_K": <(3,3)>,
        "cam_R_m2w": <(4,4)>,
        "cam_t_m2w": <(3,)>,
        "obj_id": 2
    }, ...
]}
\end{lstlisting}

\begin{lstlisting}[caption=\texttt{scene\_info.json} contains information about the data generation process for each image.\label{L5}, float]
{
    "<image id>": {
            "light": <lightmap_id>,
            "carrier": <carrier_id>,
            "parts": <0 (mix) | obj_id>,
            "viewpoint" <viewpoint_id>,
        }, ...
}
\end{lstlisting}

\begin{figure*}[t]
    \caption{Real-world image (left) and corresponding synthetic images with varying lighting conditions.\label{F9}}
    \centering
    \includegraphics[width=\textwidth]{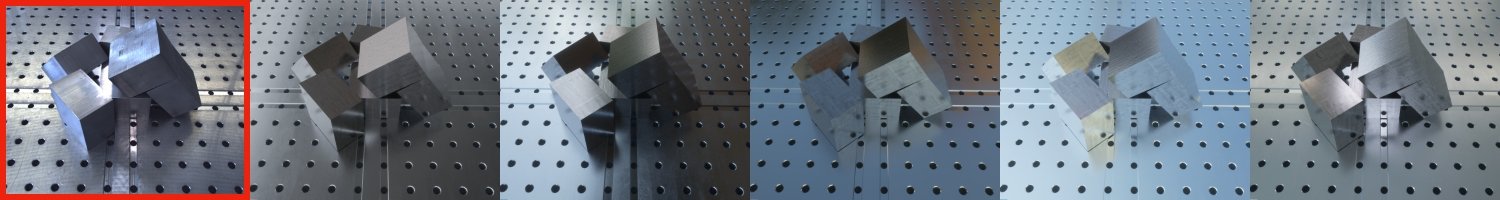}
\end{figure*}

\subsection{Project website}
The dataset is made available on the project website \url{https://pderoovere.github.io/dimo}, along with a utility script for loading data and information about the labeling tool.

\section{Conclusion}

We present a diverse dataset of industrial, reflective objects with both real-world and synthetic data. Real-world data is obtained by recording multi-view images of scenes with varying object shapes, materials, carriers, compositions and lighting conditions. This leads to over 31,200 images, accurately labelled using a new public tool. Synthetic data is obtained by carefully simulating real-world conditions and varying environmental conditions in a controlled and realistic way. This results in over 553,800 synthetic images. We hope our dataset can facilitate both sim-to-real research and research on computer vision tasks involving reflective objects.

\begin{acks}
    We thank Cedric Verschueren and Hube Van Loey for their help in annotating real-world data and RoboJob NV for their overall support.
\end{acks}

\begin{dci}
    The Authors declare that there is no conflict of interest.
\end{dci}

\begin{funding}
    The authors disclosed receipt of the following financial support for the research, authorship, and/or publication of this article: this work was supported by VLAIO Baekeland Mandate HBC.2019.2162.
\end{funding}

\bibliographystyle{SageH}
\bibliography{main}

\end{document}